\pdfoutput=1
\documentclass[twoside,11pt]{article}
\pdfoutput=1

\usepackage{jmlr2e}
\usepackage[T1]{fontenc}
\usepackage{calc}
\usepackage{amsmath}
\usepackage[capitalise]{cleveref}
\usepackage{textcomp}

\usepackage[dvipsnames]{xcolor}
\usepackage{tikz}
\usepackage{pgfplots}
\usetikzlibrary{matrix}
\usepgfplotslibrary{groupplots}
\usepgfplotslibrary{fillbetween}
\pgfplotsset{compat=newest}
\usepackage{placeins}

\usepackage{booktabs}
\usepackage{tabularx}
\usepackage{longtable}
\usepackage{threeparttable}
\usepackage{multirow}
\newcolumntype{M}[1]{>{\centering\arraybackslash}m{#1}}
\newcolumntype{L}[1]{>{\raggedright\arraybackslash}m{#1}}
\usepackage{csvsimple}
\usepackage{siunitx}

\jmlrheading{1}{}{}{}{}{}
\firstpageno{1}

\begin{document}

\title{Less Is More: A Comparison of Active Learning Strategies for 3D Medical Image Segmentation}

\author{\name Josafat-Mattias Burmeister$^{1 *}$ \email josafat-mattias.burmeister@student.hpi.de \\
        \name Marcel Fernandez Rosas$^{1 *}$ \email marcel.fernandezrosas@student.hpi.de \\
        \name Johannes Hagemann$^1$ \email johannes.hagemann@student.hpi.de \\
        \name Jonas Kordt$^1$ \email jonas.kordt@student.hpi.de \\
        \name Jasper Blum$^1$ \email jasper.blum@student.hpi.de \\
        \name Simon Shabo$^1$ \email simon.shabo@student.hpi.de \\
        \name Benjamin Bergner$^{1 \dagger}$ \email benjamin.bergner@hpi.de \\
        \name Christoph Lippert$^{1,2 \dagger}$ \email christoph.lippert@hpi.de \\
       \addr $^1$ Digital Health \& Machine Learning, Hasso Plattner Institute, University of Potsdam, Germany
       \\
       $^2$ Hasso Plattner Institute for Digital Health at Mount Sinai, Icahn School of Medicine at Mount Sinai, NYC, USA
       \\
       * equal contribution, $^\dagger$ equal advising}

\maketitle

\begin{abstract}%
Since labeling medical image data is a costly and labor-intensive process, active learning has gained much popularity in the medical image segmentation domain in recent years. A variety of active learning strategies have been proposed in the literature, but their effectiveness is highly dependent on the dataset and training scenario. To facilitate the comparison of existing strategies and provide a baseline for evaluating novel strategies, we evaluate the performance of several well-known active learning strategies on three datasets from the Medical Segmentation Decathlon. Additionally, we consider a strided sampling strategy specifically tailored to 3D image data. We demonstrate that both random and strided sampling act as strong baselines and discuss the advantages and disadvantages of the studied methods. To allow other researchers to compare their work to our results, we provide an open-source framework for benchmarking active learning strategies on a variety of medical segmentation datasets.
\end{abstract}

\begin{keywords}
  Deep Learning, Active Learning, Semantic Segmentation, Medical Imaging
\end{keywords}

\section{Introduction}

Segmentation of organs and abnormalities in medical images is important for both research and clinical applications, such as radiotherapy planning or cardiac function assessment. In recent years, it has been shown that deep learning techniques can aid in these tasks, but they require a large amount of annotated training data~\citep{lin-2021-radiotherapy,chen-2020-cardiac-segmentation}. Especially in the medical domain, data annotation is very costly and time-consuming~\citep{dacruz-2021-active-learning}. One promising approach to reducing the labor-intensive annotation process is called active learning. In this approach, the goal is to only select the most informative samples for labeling. Previous work has shown that by training on an informative subset instead of the entire dataset, the number of required labeled samples can be reduced while achieving similar performance~\citep{zhang-2019-multi-resolution,zhang-2019-sparse-annotation}. Various methods have been proposed to identify the most informative subset of a dataset. Common approaches are uncertainty sampling and representativeness sampling~\citep{budd-2021-active-learning-review}. Building on these approaches, several publications present strategies tailored to the domain of 3D medical image segmentation~\citep{top-2011-interactive-segmentation,konyushkova-2015-geometry,zhang-2019-multi-resolution,zhang-2019-sparse-annotation,zheng-2020-annotation-sparsification}. Commonly, these methods are evaluated by comparing them to random sampling, uncertainty sampling, or a model trained on the entire dataset. However, direct comparison of strategies is difficult because different authors use different datasets, model architectures, and active learning protocols. We seek to improve the comparability of active learning research in the 3D medical image segmentation domain through the following contributions:

\begin{enumerate}
    \item We establish a baseline for evaluating active learning strategies by comparing the performance of several well-known query strategies and a strided sampling strategy on three 3D medical imaging datasets. We show that both random sampling and strided sampling act as strong baselines and discuss the advantages and disadvantages of the studied methods in 3D medical image segmentation. 

    \item We introduce an open-source benchmarking framework that allows to evaluate active learning strategies using a wide range of medical segmentation datasets.\footnote{\emph{Active Segmentation framework:} \url{https://github.com/HealthML/active-segmentation}}
\end{enumerate}
\section{Related Work}
\label{sec:related-work}

In our work, we build upon previously proposed approaches to uncertainty sampling, representativeness sampling, as well as active learning with pseudo-labels.

\paragraph{Uncertainty sampling}
Approaches to capture uncertainty in a model's prediction include least confidence~\citep{sharma-2019-active-learning}, Shannon entropy~\citep{konyushkova-2015-geometry}, and prediction variance~\citep{wang-2019-two-step-query,zhang-2019-multi-resolution}. Our work evaluates the first two approaches, as they are among the best-known ones~\citep{budd-2021-active-learning-review}.

\paragraph{Representativeness sampling}
Common approaches to increase the diversity of a sampled dataset are to either select the data items that are most dissimilar to the already labeled items~\citep{sharma-2019-active-learning} or that are most similar to the unlabeled items~\citep{yang-2017-suggestive-annotation,zheng-2020-annotation-sparsification}. Image similarity is usually measured using feature vectors generated either by autoencoders~\citep{zheng-2020-annotation-sparsification} or by the prediction models~\citep{yang-2017-suggestive-annotation}. We consider the latter approach because it is more efficient in terms of training cost.

\paragraph{Active learning with pseudo-labels}
Several authors propose to include the most confident predictions of a model as pseudo-labels in the training set~\citep{gorriz-2017-cost-effective-al,zheng-2020-annotation-sparsification}. To leverage the similarity of adjacent slices in medical 3D images, we propose a strided sampling approach in which pseudo-labels are generated through interpolation.

\paragraph{Benchmarking frameworks}
In the image classification domain, several extensible active learning frameworks exist, such as modAL~\citep{danka-2018-mod-al} or ALiPy~\citep{tang-2019-alipy}. One main contribution of our work is the introduction of an active learning simulation framework for medical 3D image segmentation. Our framework is intended to be used as a benchmarking tool. It is readily extensible, experiments are comparable and reproducible.
\section{Methodology}

\subsection{Query Strategies}
\label{sec:query-strategies}

We evaluate two well-known types of active learning query strategies, uncertainty sampling and representativeness sampling. Additionally, we propose a strided sampling strategy, which exploits the three-dimensional structure of medical image data to generate pseudo-labels. Details on the query strategies are provided in \hyperref[app:query-strategies]{Appendix A}.
\begin{figure}
    \centering
    \includegraphics[width=\linewidth,trim=0cm 1cm 0cm 1cm,clip]{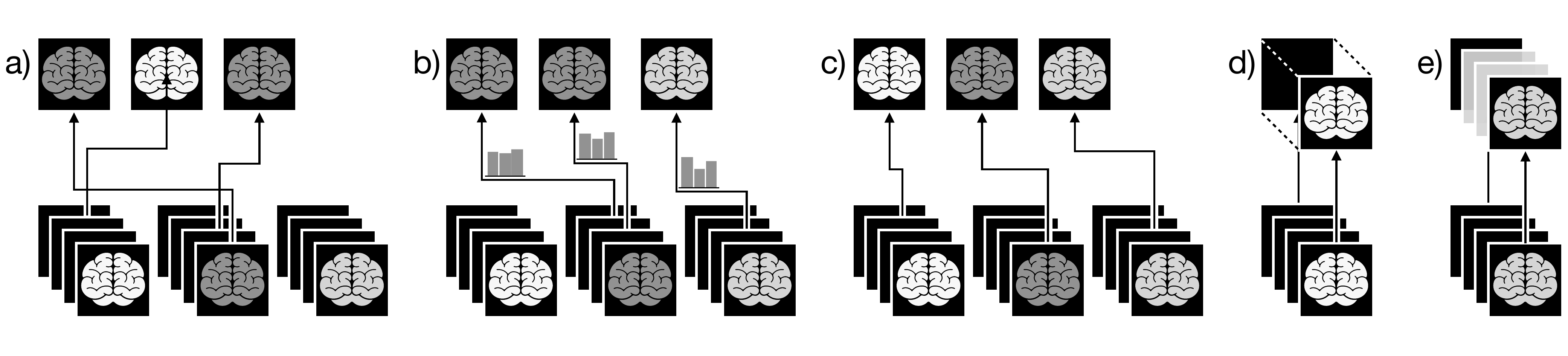}
    \caption{Visual representation of a) random sampling, b) uncertainty sampling, c) representativeness sampling, d) strided sampling, e) strided sampling with interpolation.}
    \label{fig:query-strategies}
\end{figure}
\paragraph{Uncertainty sampling} In this approach, the current model is used to generate predictions on the set of unlabeled data. The data items where the predictions are most uncertain are selected for labeling. We consider two different methods to calculate the uncertainty, least confidence (LCUS) and Shannon entropy (EntrUS).

\paragraph{Representativeness sampling} This approach aims to obtain a diverse and representative subset of a dataset. In our work, we consider two approaches that use model-generated feature vectors, namely distance-based (DistRS) and clustering-based (ClustRS) representativeness sampling. We also consider stratified random sampling (StrRS) as the most basic form of representativeness sampling, which ensures that the same number of 2D slices is sampled from each 3D scan.

\paragraph{Strided sampling \& interpolation} In this strategy, we consider sub-blocks of 3D scans, consisting of a top, a bottom, and intermediate 2D slices. Assuming that adjacent slices contain redundant information, only the top and the bottom slice of a block are selected for annotation. Intermediate slices are either not included in the training set, or pseudo-labels are generated by interpolating between the labels of the top and the bottom slice. Blocks are sampled either randomly or based on model uncertainty.

\subsection{Datasets}

We evaluate the query strategies on three tasks from the Medical Segmentation Decathlon challenge~\citep{antonelli-2021-medical-decathlon}: Heart, hippocampus, and prostate segmentation. These three tasks were selected to cover segmentation tasks with varying complexity and datasets with disparate characteristics, i.e., different modalities, image and dataset sizes. Only the training datasets of these tasks are used in our study, as no labels are publicly available for the test datasets. To validate our results on unseen data, $20\%$ of each training set was withheld for validation purposes.

\subsection{Active Learning Protocol}

We consider an active learning scenario in which 2D image slices are sampled from 3D image datasets and are annotated to train a 2D segmentation model. We prefer 2D over 3D models because initial experiments did not show a significant performance gap and the higher training speed of 2D models is better suited for interactive active learning scenarios. To train the initial models, 32 image slices are randomly sampled and annotated. Subsequently, 50 active learning iterations are performed. Each active learning iteration consists of a simulated labeling step and a training step. In the labeling step, 16 image slices are selected using a query strategy and annotated. In the training step, the model is trained for 10 epochs on the enlarged training dataset. To incrementally fit the model to the sampled data, we employ a fine-tuning approach in which model weights are not reset between active learning iterations. In all experiments, we use the 2D U-Net architecture \citep{ronneberger-2015-u-net}. Based on initial experiments, we use the  Adam optimizer~\citep{kingma-2017-adam} with a fixed learning rate of $10^{-4}$ and focal loss with $\gamma = 5$. Depending on image size, we choose batch sizes between 16 and 264. To ensure full reproducibility of our results, all random processes, such as model initialization or data shuffling, are seeded. Each experiment is repeated three times with different seeds. We choose the average and standard deviation of the mean validation Dice scores of these three runs as evaluation metrics. Details on our active learning framework are provided in \hyperref[app:al-framework]{Appendix B}.

\section{Results}
\label{sec:results}

\begin{figure}
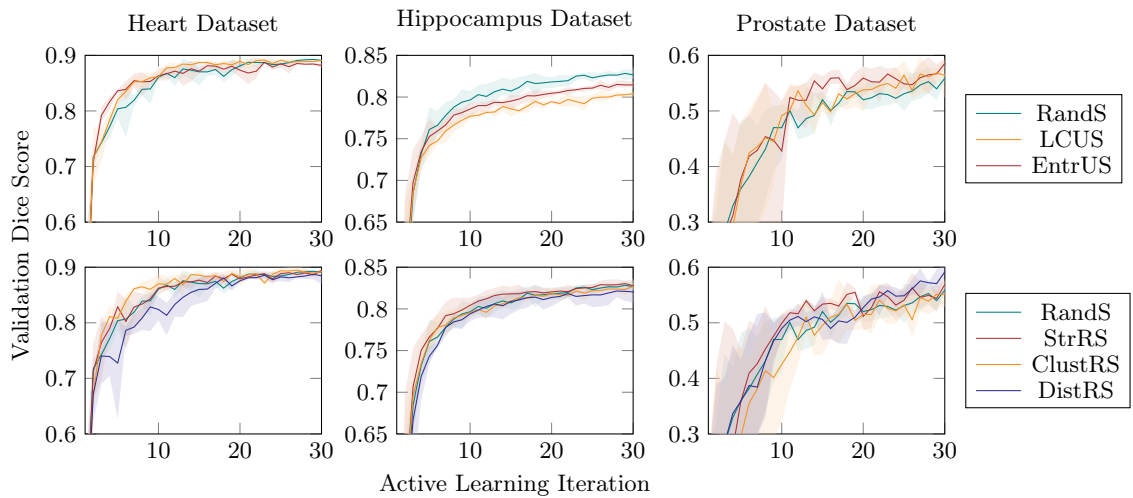

\begin{tikzpicture}
\footnotesize
\begin{groupplot}[group style={
	group name=baseline-sampling,
	group size= 3 by 2,
	vertical sep=0.6cm
	},
	height=3.8cm,width=0.31\linewidth]
\input{plots/plot_uncertainty_sampling}
\input{plots/plots_representativeness_sampling}
\end{groupplot}
%
\matrix[
    matrix of nodes,
    anchor=west,
    draw,
    inner sep=0.2em,
    draw
  ]at([xshift=2ex]baseline-sampling c3r1.east)
  {
    \ref{plots:random-sampling-uncertainty}& RandS \\
    \ref{plots:uncertainty-sampling-least-confidence}& LCUS\\
    \ref{plots:uncertainty-sampling-entropy}& EntrUS \\
  };
\matrix[
    matrix of nodes,
    anchor=west,
    draw,
    inner sep=0.2em,
    draw
  ]at([xshift=2ex]baseline-sampling c3r2.east)
  {
    \ref{plots:random-sampling-representativeness}& RandS \\
    \ref{plots:representativness-sampling-stratified}& StrRS \\
    \ref{plots:representativness-sampling-clustering}& ClustRS \\
    \ref{plots:representativness-sampling-distance}& DistRS\\
  };
\end{tikzpicture}
\caption{Performance of uncertainty sampling (first row) and representativeness sampling (second row) compared to a random sampling baseline.}
\label{fig:baseline-strategies}
\end{figure}

\begin{figure}
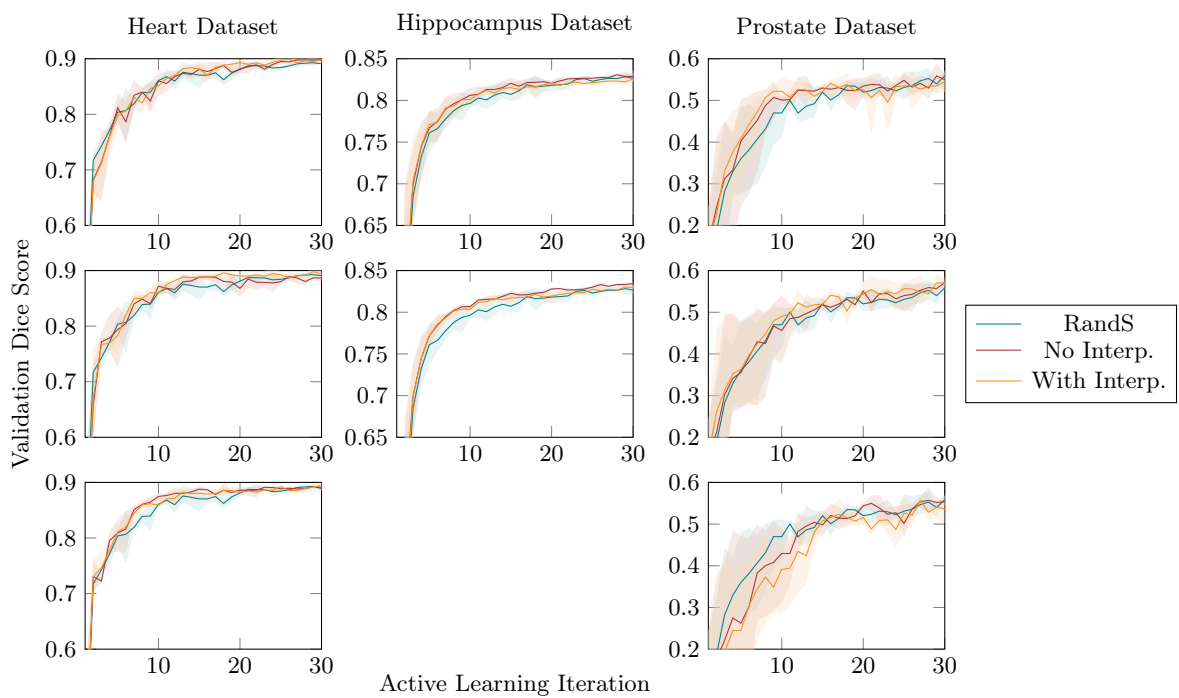

\begin{tikzpicture}
\footnotesize
\begin{groupplot}[group style={
	group name=block-interpolation-sampling,
	group size= 3 by 3,
	vertical sep=0.6cm
	},
	height=3.8cm,width=0.31\linewidth]
\input{plots/plots_interpolation_5}
\input{plots/plots_interpolation_10}
\input{plots/plots_interpolation_15}
\end{groupplot}
%
\matrix[
    matrix of nodes,
    anchor=west,
    draw,
    inner sep=0.2em,
    draw
  ]at([xshift=2ex]block-interpolation-sampling c3r2.east)
  {
    \ref{plots:random-sampling-interpolation}& RandS \\
    \ref{plots:random-block-selection}& No Interp. \\
    \ref{plots:random-interpolation-signed-distance}& With Interp. \\
    };
\end{tikzpicture}
\caption{Performance of strided sampling with block size 5 (first row), 10 (second row), and 15 (third row). The results shown were obtained with random block sampling and signed distance interpolation. For the hippocampus dataset, block size 15 was not tested since all scans in this dataset contain less than 15 slices.}
\label{fig:custom-strategies}
\end{figure}

\paragraph{Baseline strategies} As shown in \cref{fig:baseline-strategies} and \cref{fig:custom-strategies}, most query strategies performed similarly on the three datasets studied, and no strategy outperformed random sampling (RandS) by a large margin. Which strategy worked best was highly dataset dependent: On the heart dataset, uncertainty sampling strategies slightly outperformed RandS in the early active learning iterations. DistRS slowed down the learning process significantly, while ClustRS produced a steeper learning curve than RandS. On the hippocampus dataset, both uncertainty sampling strategies resulted in flatter learning curves than RandS. All representativeness sampling strategies performed similarly. On the prostate dataset, EntrUS outperformed RandS in later iterations. StrRS also slightly outperformed RandS, whereas ClustRS sampling did not perform well.

\paragraph{Strided sampling \& interpolation} In this section, we consider random sampling-based strided sampling (StrideS) with and without interpolation. Results for further variants of strided sampling are provided in \hyperref[app:results]{Appendix C}. We consider block sizes 5, 10, and 15. StrideS without interpolation outperformed the RandS baseline for some combinations of block size and dataset, showing that it is already effective without generating pseudo-labels. On the hippocampus dataset, StrideS without interpolation exceeded RandS for all tested block sizes. On the heart dataset, StrideS outperformed RandS for block size $15$, while on the prostate dataset, block size $5$ produced the best results. Comparing StrideS with and without the generation of pseudo-labels showed that the inclusion of pseudo-labels barely improved model performance in most cases. For all three datasets, pseudo-label quality decreased with increasing block size. For the prostate dataset, e.g., the average interpolation Dice score was $0.88 \pm 0.12$ at block size $5$, while it decreased to $0.81 \pm 0.17$ at block size $15$.
\section{Discussion}
\label{sec:discussion}

Our results show that the performance of uncertainty sampling and model-driven representativeness sampling strategies varies considerably between datasets, while stratified random sampling and strided sampling yield decent results on all tested datasets. However, none of the tested strategies outperformed the random sampling baseline by a large margin. Our results are thus in line with the findings of \citeauthor{nath-2021-image-segmentation}, who state that random sampling is a surprisingly difficult baseline to beat and that working active learning solutions are highly dataset dependent~\citep{nath-2021-image-segmentation}. In the following, we discuss reasons for this and consider the strengths of simple strategies such as stratified and strided sampling.

\paragraph{Uncertainty sampling} The idea to select samples the model is most uncertain about is based on the idealized assumption that all samples contain independent, equally relevant information. When sampling 2D slices from 3D images, uncertainty sampling may result in selecting non-representative or redundant slices. To address this issue, a combination with representativeness sampling techniques might be beneficial. Another concern with uncertainty sampling is that neural networks tend to be overconfident, meaning that a model's predictions may not reflect its true uncertainty~\citep{guo-2017-calibration}. Several approaches to this problem exist, such as model calibration~\citep{wang-2021-calibration}, Bootstrapping~\citep{yang-2017-suggestive-annotation}, or the use of Bayesian models~\citep{maddox-2019-bayesian-networks}, but not all have been studied in the context of 3D medical image segmentation.

\paragraph{Representativeness sampling} Model-driven representativeness strategies rely on model-generated feature vectors. At an early stage of active learning, however, this feature representation might not be meaningful as the model is trained on very few samples initially. This issue could be addressed by pre-training models in a self-supervised manner. Model-independent strategies such as stratified random sampling can also be an alternative, as our results show.

\paragraph{Strided sampling \& interpolation} When appropriate block sizes were chosen, strided sampling outperformed random sampling on all datasets. Thus, it can act as a strong baseline for 3D medical image segmentation. Strided sampling has one strength in common with stratified random sampling: both strategies evenly sample 2D slices from the 3D scans ensuring that the entire dataset is covered. Additionally, strided sampling avoids the selection of neighboring, redundant slices. A trade-off must be considered when choosing the block size. Although the variance of the sampled data increases with larger blocks, resulting in a larger and more diverse dataset, the quality of the pseudo-labels decreases.

\section{Conclusion}

By comparing several basic active learning strategies, we have established a baseline for evaluating active learning strategies in the 3D medical image segmentation domain. Our results show that in the considered scenario, simple approaches such as random sampling or strided sampling are robust and may be sufficient for many use cases. Uncertainty sampling and model-driven representativeness sampling strategies seem to require more fine-tuning and possibly the integration of expert knowledge on dataset characteristics. By open-sourcing our evaluation framework, we aim to stimulate further research in this area.

\bibliography{paper}

\newpage

\appendix
\section*{Appendix A.}
\label{app:query-strategies}

In this supplementary material, implementation details of all query strategies evaluated in this work are provided.

\subsection*{Uncertainty Sampling}

In uncertainty sampling, the data items where a model's predictions are most uncertain are selected for labeling. We consider two well-known methods for capturing model uncertainty: least confidence~\citep{sharma-2019-active-learning} and Shannon entropy~\citep{shannon-1948-entropy,top-2011-interactive-segmentation,konyushkova-2015-geometry}.

\paragraph{Least confidence-based sampling} In least confidence-based sampling, uncertainty is computed as follows: For each slice $x \in \mathbb{R}^{n\times m}$ of each 3D scan from the unlabeled pool the prediction $P_{\theta}(y|x)$ under the current model $\theta$ is computed, where $n$ is the image height and $m$ is the image width. For each pixel, the distance of the prediction to a maximum uncertainty value $\alpha$ is computed. The closer the predicted probability is to $\alpha$, the more uncertain the model. Then, the uncertainty of a slice $x$ is given as the negative sum of the class- and pixel-wise distance measures:

\begin{equation*}
\label{equ:uncertainty-distance}
    x_{LC}^*(x) = - \sum_{c=1}^C\sum_{i=1}^n \sum_{j=1}^m \lvert \alpha - P_{\theta}(y_{c,i,j}|x) \rvert
\end{equation*}

with

\begin{equation*}
\label{equ:uncertainty-alpha-1}
\alpha = \left\{
         \begin{array}{ll}
            0.5 & \text{in multi-label scenario} \\
            \frac{1}{C} & \text{in single-label scenario with } C=\#classes \\
         \end{array}
    \right.
\end{equation*}

\paragraph{Entropy-based sampling} Another approach to measure uncertainty is Shannon entropy. In this approach, the uncertainty of a slice $x \in \mathbb{R}^{n\times m}$ is calculated as the sum of the pixel-wise distance measures multiplied by the logarithm of the pixel-wise distance measures:

\begin{equation*}
\label{equ:uncertainty-entropy}
    x_{H}^*(x) = \sum_{c=1}^C\sum_{i=1}^n \sum_{j=1}^m \lvert \alpha - P_{\theta}(y_{c,i,j}|x) \rvert \times \log_{e} \lvert \alpha - P_{\theta}(y_{c,i,j}|x) \rvert 
\end{equation*}

with

\begin{equation*}
\label{equ:uncertainty-alpha-2}
\alpha = \left\{
         \begin{array}{ll}
            0.5 & \text{in multi-label scenario} \\
            \frac{1}{C} & \text{in single-label scenario with } C=\#classes \\
         \end{array}
    \right.
\end{equation*}

\noindent Since the uncertainty of a model is usually similar for adjacent slices of a 3D scan, uncertainty sampling can result in multiple redundant slices of the same 3D scan being sampled. To mitigate this effect, we limit uncertainty sampling to one slice per 3D scan in each active learning iteration unless the number of 3D scans is less than the number of slices to be sampled.

\newpage
\subsection*{Representativeness Sampling}

Representativeness sampling aims to increase the diversity and representativeness of the sampled training set. In this work, we study one model-independent representativeness sampling strategy, stratified random sampling, and two model-driven strategies, clustering-based sampling and distance-based sampling.

\paragraph{Stratified random sampling} In this most basic type of representativeness sampling, slices are sampled randomly from the unlabeled 3D scans, while ensuring that the same number of slices is sampled from each 3D scan (each 3D scan is considered a stratum).
\\
\\
Model-driven strategies measure image similarity using model-generated feature vectors that represent the content of an image slice in a compressed form. To obtain such feature vectors, for each unlabeled slice $x \in \mathbb{R}^{n\times m}$ of each 3D scan, the prediction under the current model $\theta$ is computed, and a feature vector is retrieved from one of the inner model layers. In our implementation, the bottleneck layer of the U-Net architecture is used to retrieve feature vectors. To reduce the dimensionality of the obtained feature vectors, spatial max-pooling is applied, followed by a principal component analysis (PCA). In our implementation, the spatial dimension is aggregated into a single value by spatial max-pooling, and then the feature channels are compressed to ten principal components.

\paragraph{Clustering-based representativeness sampling} In the clustering-based approach, we apply mean shift clustering with a fixed bandwidth of $5$ to the feature vectors. Unlabeled slices are selected from the obtained clusters such that each cluster is equally represented in the training set relative to its size.

\paragraph{Distance-based representativeness sampling} In the distance-based approach, the average Euclidean distance between each feature vector of the set of unlabeled slices and each feature vector of the training set is computed:

\begin{equation*}
\label{equ:distance-sampling}
    s(x) = \frac{1}{N} \sum_{i=1}^N d(f_i, f_x)
\end{equation*}

{\noindent}where $N$ is the training set size, $f_i$ is the feature vector of the $i$-th slice in the training set, $f_x$ is the feature vector of an unlabeled slice $x$, and $d$ is the Euclidean distance. The unlabeled slices with the largest average distance to the feature vectors from the training set are selected as the most representative. A similar sampling technique is described in the work of \cite{smailagic-2018-medal}.

\newpage
\subsection*{Strided Sampling}

To combine the computational efficiency of a 2D model with the three-dimensional structure of medical image data, we propose a strided sampling strategy that generates pseudo-labels through interpolation. Since adjacent slices of a 3D scan contain redundant information, strided sampling extracts sub-blocks of 3D scans, consisting of a top, a bottom, and intermediate 2D slices, and selects only the top and the bottom slice for annotation. The intermediate slices are either not included in the training set, or pseudo-labels are generated by interpolating between the labels of the top and the bottom slice. The first case, strided sampling without interpolation, is equivalent to the equal interval query strategy proposed in the work of \cite{zhang-2019-sparse-annotation}. The second case, strided sampling with pseudo-labels, is similar to the works of \cite{gorriz-2017-cost-effective-al} and \cite{zheng-2020-annotation-sparsification}, who also make use of pseudo-labels. In our case, however, the pseudo-labels are not generated by a model but by interpolation, meaning that their quality is independent of the quality of the model. We experiment with two different interpolation methods, namely signed distance interpolation and morphological contour interpolation. 

\paragraph{Signed distance interpolation}

This interpolation method computes distance maps for the top and bottom slices of a selected block and is inspired by a Stackoverflow question.\footnote{\emph{Interpolation between two images:} \href{https://stackoverflow.com/questions/48818373/interpolate-between-two-images}{stackoverflow.com/questions/48818373/interpolate-between-two-images}} These distance maps are calculated separately for each class channel and indicate the signed distance to the edge of the segmentation mask for each pixel $x_{i, j}$ in a two-dimensional image slice $x \in \mathbb{R}^{n\times m}$. Pixels belonging to the foreground class are assigned a positive distance value and pixels belonging to the background class a negative distance value. To calculate the distance maps, the signed distance function $d(x)$ is used, which is defined as

\begin{equation*}
\label{equ:signed-distance}
    d(x)=edt(x) - edt({x^{-1}}) + 0.5
\end{equation*}

{\noindent}where $x^{-1}$ is the inverted segmentation mask and 
\begin{equation*}
\label{equ:euc-distance}
    edt(x)_{i,j} = \| x_{i,j} - b_{i,j} \|
\end{equation*}

{\noindent}is the Euclidian distance transform where $b_{i,j}$ is the background pixel with the smallest Euclidean distance to input pixel $x_{i,j}$.
\\
To obtain an interpolated segmentation mask, we interpolate between the distance maps of the top and bottom slices. All pixels with a positive distance value in the interpolated distance map are assigned to the foreground class and all other pixels to the background class.

\begin{figure}
    \centering
    \includegraphics[width=12cm]{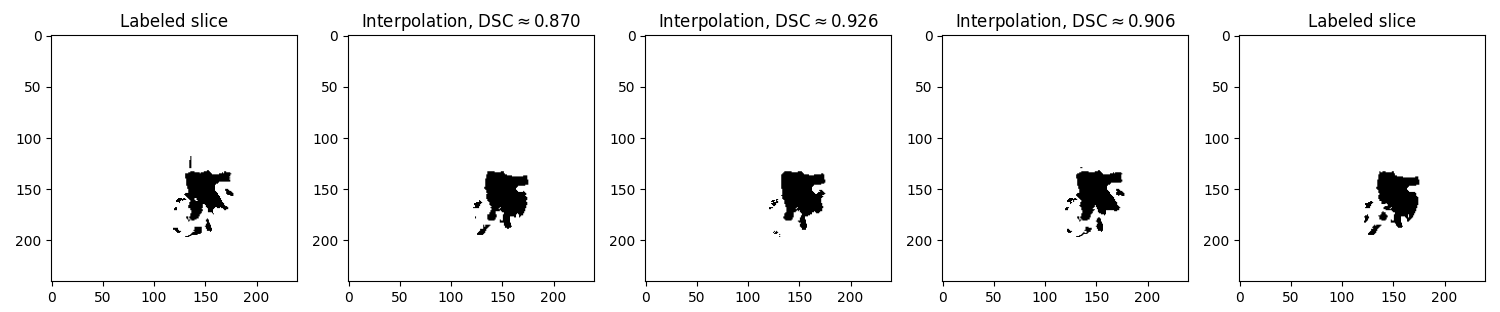}
    \caption{Example interpolation with signed distance method of a block of a brain tumor segmentation mask with block size $=5$.}
    \label{fig:signed-distance-example}
\end{figure}

\paragraph{Morphological contour interpolation}

For the morphological contour interpolation, we use an implementation of the Insight Toolkit (ITK) introduced by \cite{zukic-2016-itk}. The approach is based on decomposing a many-to-many correspondence into three fundamental cases: one-to-one, one-to-many, and zero-to-one correspondences~\citep{albu-2008-morph-interpolation}.
\\
\begin{figure}
    \centering
    \includegraphics[width=12cm]{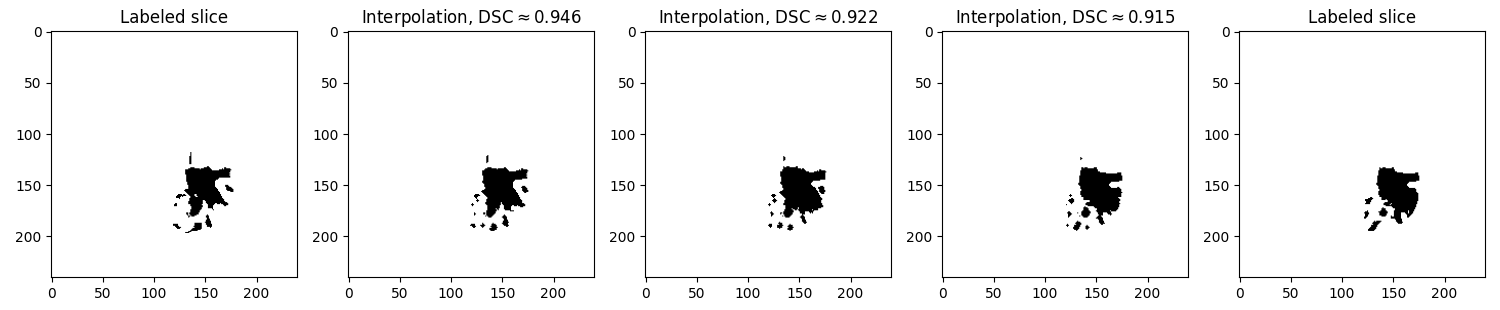}
    \caption{Example interpolation with morphological contour method of a block of a brain tumor segmentation mask with block size $=5$.}
    \label{fig:morph-contour-example}
\end{figure}
\\
\\
\\
{\noindent}If the segmentation masks of the top and bottom slice of a selected block do not overlap, no interpolation of the segmentation masks is possible. In such cases, no pseudo-labels are created and only the upper and lower slices of the block are added to the training set.
\\
\\
The sampling of blocks in strided sampling can be done either randomly, or according to another criterion, e.g. uncertainty. The distance between the top and bottom slice of a block is determined by the hyperparameter~$l$, the maximum block size. If labels or pseudo-labels already exist for some slices of a selected block, we discard it and sample another, completely unlabeled block instead. If there are no more fully unlabeled blocks for the given block size~$l$, the block size is reduced before block sampling continues. Once all slices in the dataset have been assigned either a label or a pseudo-label, the block size is reset to the original block size~$l$, and block selection subsequently also accepts blocks with pseudo-labels, so that pseudo-labels are gradually replaced by true labels.

\newpage
\section*{Appendix B.}
\label{app:al-framework}

To compare different query strategies, we developed an active learning simulation framework that allows evaluating active learning strategies on different medical image datasets. The framework provides two model training modes: classic training on a fully labeled dataset and active learning. Both modes are compatible with two-dimensional segmentation models, and the classic mode additionally supports three-dimensional models. The framework comes with eleven ready-to-use datasets, and custom datasets can be integrated with little effort. The provided datasets include the two-dimensional Breast Cancer Semantic Segmentation dataset~\citep{amgad-2019-bcss} and ten three-dimensional datasets from the Medical Segmentation Decathlon challenge~\citep{antonelli-2021-medical-decathlon}. In the active learning mode, the framework implements the query strategies described in \cref{sec:query-strategies}. Since the framework focuses on benchmarking and comparing different query strategies, it provides a reproducibility mode that enables different active learning strategies to be executed under the same conditions. For this purpose, all random processes, such as model and training pool initialization, are seeded, and only deterministic operations are used. In addition, the framework allows tracking various metrics and losses such as the Dice score, Hausdorff distance, sensitivity, and specificity. Several loss functions are implemented, including Dice loss~\citep{milletari-2016-dice-loss}, cross-entropy loss, and focal loss~\citep{lin-2017-focal-loss}. The framework is implemented using PyTorch\footnote{\emph{PyTorch framework:} \url{https://pytorch.org}}~\citep{paszke-2019-pytorch} and PyTorch Lightning~\citep{falcon-2019-pytorch-lightning}.\footnote{\emph{PyTorch Lightning framework:} \url{https://www.pytorchlightning.ai}} Due to its modular structure, it can be easily extended with additional datasets, model architectures, query strategies, and evaluation metrics. More details can be found in the framework repository on GitHub\footnote{\emph{Active Segmentation framework:} \url{https://github.com/HealthML/active-segmentation}} and in the framework documentation.\footnote{\emph{Active Segmentation framework documentation:} \url{https://healthml.github.io/active-segmentation/}}

\newpage
\section*{Appendix C.}
\label{app:results}
\newcommand{\strategyName}[3]{%
  \ifthenelse{\equal{#1}{Random sampling}}
    {#1}
    {#2{, }#3}
}
In this supplementary material, we provide the mean validation Dice scores of the foreground classes for all tested active learning strategies for selected active learning iterations. The following tables show the average and standard deviation of the mean validation Dice scores from three experiment runs.
\begin{table}[ht]
    \footnotesize
    \centering
    \caption{Performance of single slice query strategies on the heart dataset.}
    \label{table:results-baseline-strategies-heart}
        \csvloop{
            file=data/heart/baseline_strategies.csv,
            head to column names,
            tabular={
                m{0.3\textwidth-2\tabcolsep}
                S[table-format=1.3(3), separate-uncertainty, table-column-width=0.135\textwidth-2\tabcolsep]
                S[table-format=1.3(3), separate-uncertainty, table-column-width=0.135\textwidth-2\tabcolsep]
                S[table-format=1.3(3), separate-uncertainty, table-column-width=0.135\textwidth-2\tabcolsep]
                S[table-format=1.3(3), separate-uncertainty, table-column-width=0.135\textwidth-2\tabcolsep]
                S[table-format=1.3(3), separate-uncertainty, table-column-width=0.135\textwidth-2\tabcolsep]
                c@{}
            },
            table head=\toprule\multirow{3}{*}{\parbox{0.45\textwidth-2\tabcolsep}{Query  Strategy}} & \multicolumn{5}{c}{Active Learning Iteration} \\ \\ & {5} & {10} & {15} & {20} & {50} \\ \midrule,
            command=
            \Strategy &
            \DiceScoreMeanFive\pm\DiceScoreStdFive &
            \DiceScoreMeanTen\pm\DiceScoreStdTen &
            \DiceScoreMeanFiveteen\pm\DiceScoreStdFiveteen &
            \DiceScoreMeanTwenty\pm\DiceScoreStdTwenty &
            \DiceScoreMeanFifty\pm\DiceScoreStdFifty &
            \empty,
            table foot=\bottomrule}
\end{table}
\begin{table}[ht]
    \footnotesize
    \centering
    \caption{Performance of single slice query strategies on the hippocampus dataset.}
    \label{table:results-baseline-strategies-hippocampus}
        \csvloop{
            file=data/hippocampus/baseline_strategies.csv,
            head to column names,
            tabular={
                m{0.3\textwidth-2\tabcolsep}
                S[table-format=1.3(3), separate-uncertainty, table-column-width=0.135\textwidth-2\tabcolsep]
                S[table-format=1.3(3), separate-uncertainty, table-column-width=0.135\textwidth-2\tabcolsep]
                S[table-format=1.3(3), separate-uncertainty, table-column-width=0.135\textwidth-2\tabcolsep]
                S[table-format=1.3(3), separate-uncertainty, table-column-width=0.135\textwidth-2\tabcolsep]
                S[table-format=1.3(3), separate-uncertainty, table-column-width=0.135\textwidth-2\tabcolsep]
                c@{}
            },
            table head=\toprule\multirow{3}{*}{\parbox{0.45\textwidth-2\tabcolsep}{Query  Strategy}} & \multicolumn{5}{c}{Active Learning Iteration} \\ \\ & {5} & {10} & {15} & {20} & {50} \\ \midrule,
            command=
            \Strategy &
            \DiceScoreMeanFive\pm\DiceScoreStdFive &
            \DiceScoreMeanTen\pm\DiceScoreStdTen &
            \DiceScoreMeanFiveteen\pm\DiceScoreStdFiveteen &
            \DiceScoreMeanTwenty\pm\DiceScoreStdTwenty &
            \DiceScoreMeanFifty\pm\DiceScoreStdFifty &
            \empty,
            table foot=\bottomrule}
\end{table}
\begin{table}[ht]
    \footnotesize
    \centering
    \caption{Performance of single slice query strategies on the prostate dataset.}
    \label{table:results-baseline-strategies-prostate}
        \csvloop{
            file=data/prostate/baseline_strategies.csv,
            head to column names,
            tabular={
                m{0.3\textwidth-2\tabcolsep}
                S[table-format=1.3(3), separate-uncertainty, table-column-width=0.135\textwidth-2\tabcolsep]
                S[table-format=1.3(3), separate-uncertainty, table-column-width=0.135\textwidth-2\tabcolsep]
                S[table-format=1.3(3), separate-uncertainty, table-column-width=0.135\textwidth-2\tabcolsep]
                S[table-format=1.3(3), separate-uncertainty, table-column-width=0.135\textwidth-2\tabcolsep]
                S[table-format=1.3(3), separate-uncertainty, table-column-width=0.135\textwidth-2\tabcolsep]
                c@{}
            },
            table head=\toprule\multirow{3}{*}{\parbox{0.45\textwidth-2\tabcolsep}{Query  Strategy}} & \multicolumn{5}{c}{Active Learning Iteration} \\ \\ & {5} & {10} & {15} & {20} & {50} \\ \midrule,
            command=
            \Strategy &
            \DiceScoreMeanFive\pm\DiceScoreStdFive &
            \DiceScoreMeanTen\pm\DiceScoreStdTen &
            \DiceScoreMeanFiveteen\pm\DiceScoreStdFiveteen &
            \DiceScoreMeanTwenty\pm\DiceScoreStdTwenty &
            \DiceScoreMeanFifty\pm\DiceScoreStdFifty &
            \empty,
            table foot=\bottomrule}
\end{table}
\FloatBarrier

{\noindent}For the strided sampling strategy, we experimented with two approaches for block selection, namely random sampling and entropy-based uncertainty sampling. For both approaches, we tested two interpolation methods, signed distance interpolation and morphological contour interpolation. In \cref{sec:results} and \cref{sec:discussion}, we focus on the results obtained with random block selection and signed distance interpolation, as signed distance interpolation outperformed morphological contour interpolation (\cref{table:results-interpolation-quality}) and random sampling produced decent results on all datasets. Uncertainty-based strided sampling performed worse than the random sampling baseline on the hippocampus dataset, while in some cases, e.g., on the heart dataset, it performed slightly better. In \cref{sec:results} and \cref{sec:discussion}, the term "strided sampling" always refers to random sampling-based strided sampling with signed distance interpolation. In the following, we provide the results of all tested variants.
\begin{table}[ht]
    \footnotesize
    \centering
    \caption{Performance of random sampling-based strided sampling strategies on the heart dataset.}
    \label{table:results-random-strided-sampling-heart}
        \csvloop{
            file=data/heart/random_strided_sampling.csv,
            head to column names,
            tabular={
                m{0.3\textwidth-2\tabcolsep}
                S[table-format=1.3(3), separate-uncertainty, table-column-width=0.135\textwidth-2\tabcolsep]
                S[table-format=1.3(3), separate-uncertainty, table-column-width=0.135\textwidth-2\tabcolsep]
                S[table-format=1.3(3), separate-uncertainty, table-column-width=0.135\textwidth-2\tabcolsep]
                S[table-format=1.3(3), separate-uncertainty, table-column-width=0.135\textwidth-2\tabcolsep]
                S[table-format=1.3(3), separate-uncertainty, table-column-width=0.135\textwidth-2\tabcolsep]
                c@{}
            },
            table head=\toprule \multirow{3}{*}{\parbox{0.45\textwidth-2\tabcolsep}{Query  Strategy}} & \multicolumn{5}{c}{Active Learning Iteration} \\ \\ & {5} & {10} & {15} & {20} & {50} \\ \midrule,
            command=
            \strategyName{\Strategy}{\Interpolation}{\BlockSize} &
            \meanDiceScoreMeanFive\pm\meanDiceScoreStdFive &
            \meanDiceScoreMeanTen\pm\meanDiceScoreStdTen &
            \meanDiceScoreMeanFiveteen\pm\meanDiceScoreStdFiveteen &
            \meanDiceScoreMeanTwenty\pm\meanDiceScoreStdTwenty &
            \meanDiceScoreMeanFifty\pm\meanDiceScoreStdFifty &
            \empty,
            table foot=\bottomrule}
\end{table}
\begin{table}[ht]
    \footnotesize
    \centering
    \caption{Performance of random sampling-based strided sampling strategies on the hippocampus dataset. For the hippocampus dataset, block size 15 was not tested since all scans in this dataset contain less than 15 slices.}
    \label{table:results-random-strided-sampling-hippocampus}
        \csvloop{
            file=data/hippocampus/random_strided_sampling.csv,
            head to column names,
            tabular={
                m{0.3\textwidth-2\tabcolsep}
                S[table-format=1.3(3), separate-uncertainty, table-column-width=0.135\textwidth-2\tabcolsep]
                S[table-format=1.3(3), separate-uncertainty, table-column-width=0.135\textwidth-2\tabcolsep]
                S[table-format=1.3(3), separate-uncertainty, table-column-width=0.135\textwidth-2\tabcolsep]
                S[table-format=1.3(3), separate-uncertainty, table-column-width=0.135\textwidth-2\tabcolsep]
                S[table-format=1.3(3), separate-uncertainty, table-column-width=0.135\textwidth-2\tabcolsep]
                c@{}
            },
            table head=\toprule \multirow{3}{*}{\parbox{0.45\textwidth-2\tabcolsep}{Query  Strategy}} & \multicolumn{5}{c}{Active Learning Iteration} \\ \\ & {5} & {10} & {15} & {20} & {50} \\ \midrule,
            command=
            \strategyName{\Strategy}{\Interpolation}{\BlockSize} &
            \meanDiceScoreMeanFive\pm\meanDiceScoreStdFive &
            \meanDiceScoreMeanTen\pm\meanDiceScoreStdTen &
            \meanDiceScoreMeanFiveteen\pm\meanDiceScoreStdFiveteen &
            \meanDiceScoreMeanTwenty\pm\meanDiceScoreStdTwenty &
            \meanDiceScoreMeanFifty\pm\meanDiceScoreStdFifty &
            \empty,
            table foot=\bottomrule}
\end{table}
\begin{table}[ht]
    \footnotesize
    \centering
    \caption{Performance of random sampling-based strided sampling strategies on the prostate dataset.}
    \label{table:results-random-strided-sampling-prostate}
        \csvloop{
            file=data/prostate/random_strided_sampling.csv,
            head to column names,
            tabular={
                m{0.3\textwidth-2\tabcolsep}
                S[table-format=1.3(3), separate-uncertainty, table-column-width=0.135\textwidth-2\tabcolsep]
                S[table-format=1.3(3), separate-uncertainty, table-column-width=0.135\textwidth-2\tabcolsep]
                S[table-format=1.3(3), separate-uncertainty, table-column-width=0.135\textwidth-2\tabcolsep]
                S[table-format=1.3(3), separate-uncertainty, table-column-width=0.135\textwidth-2\tabcolsep]
                S[table-format=1.3(3), separate-uncertainty, table-column-width=0.135\textwidth-2\tabcolsep]
                c@{}
            },
            table head=\toprule \multirow{3}{*}{\parbox{0.45\textwidth-2\tabcolsep}{Query  Strategy}} & \multicolumn{5}{c}{Active Learning Iteration} \\ \\ & {5} & {10} & {15} & {20} & {50} \\ \midrule,
            command=
            \strategyName{\Strategy}{\Interpolation}{\BlockSize} &
            \meanDiceScoreMeanFive\pm\meanDiceScoreStdFive &
            \meanDiceScoreMeanTen\pm\meanDiceScoreStdTen &
            \meanDiceScoreMeanFiveteen\pm\meanDiceScoreStdFiveteen &
            \meanDiceScoreMeanTwenty\pm\meanDiceScoreStdTwenty &
            \meanDiceScoreMeanFifty\pm\meanDiceScoreStdFifty &
            \empty,
            table foot=\bottomrule}
\end{table}
\begin{table}[ht]
    \footnotesize
    \centering
    \caption{Performance of uncertainty sampling-based strided sampling strategies on the heart dataset.}
    \label{table:results-uncertainty-strided-sampling-heart}
        \csvloop{
            file=data/heart/uncertainty_strided_sampling.csv,
            head to column names,
            tabular={
                m{0.3\textwidth-2\tabcolsep}
                S[table-format=1.3(3), separate-uncertainty, table-column-width=0.135\textwidth-2\tabcolsep]
                S[table-format=1.3(3), separate-uncertainty, table-column-width=0.135\textwidth-2\tabcolsep]
                S[table-format=1.3(3), separate-uncertainty, table-column-width=0.135\textwidth-2\tabcolsep]
                S[table-format=1.3(3), separate-uncertainty, table-column-width=0.135\textwidth-2\tabcolsep]
                S[table-format=1.3(3), separate-uncertainty, table-column-width=0.135\textwidth-2\tabcolsep]
                c@{}
            },
            table head=\toprule \multirow{3}{*}{\parbox{0.45\textwidth-2\tabcolsep}{Query  Strategy}} & \multicolumn{5}{c}{Active Learning Iteration} \\ \\ & {5} & {10} & {15} & {20} & {50} \\ \midrule,
            command=
            \strategyName{\Strategy}{\Interpolation}{\BlockSize} &
            \meanDiceScoreMeanFive\pm\meanDiceScoreStdFive &
            \meanDiceScoreMeanTen\pm\meanDiceScoreStdTen &
            \meanDiceScoreMeanFiveteen\pm\meanDiceScoreStdFiveteen &
            \meanDiceScoreMeanTwenty\pm\meanDiceScoreStdTwenty &
            \meanDiceScoreMeanFifty\pm\meanDiceScoreStdFifty &
            \empty,
            table foot=\bottomrule}
\end{table}
\begin{table}[ht]
    \footnotesize
    \centering
    \caption{Performance of uncertainty sampling-based strided sampling strategies on the hippocampus dataset. For the hippocampus dataset, block size 15 was not tested since all scans in this dataset contain less than 15 slices.}
    \label{table:results-uncertainty-strided-sampling-hippocampus}
        \csvloop{
            file=data/hippocampus/uncertainty_strided_sampling.csv,
            head to column names,
            tabular={
                m{0.3\textwidth-2\tabcolsep}
                S[table-format=1.3(3), separate-uncertainty, table-column-width=0.135\textwidth-2\tabcolsep]
                S[table-format=1.3(3), separate-uncertainty, table-column-width=0.135\textwidth-2\tabcolsep]
                S[table-format=1.3(3), separate-uncertainty, table-column-width=0.135\textwidth-2\tabcolsep]
                S[table-format=1.3(3), separate-uncertainty, table-column-width=0.135\textwidth-2\tabcolsep]
                S[table-format=1.3(3), separate-uncertainty, table-column-width=0.135\textwidth-2\tabcolsep]
                c@{}
            },
            table head=\toprule \multirow{3}{*}{\parbox{0.45\textwidth-2\tabcolsep}{Query  Strategy}} & \multicolumn{5}{c}{Active Learning Iteration} \\ \\ & {5} & {10} & {15} & {20} & {50} \\ \midrule,
            command=
            \strategyName{\Strategy}{\Interpolation}{\BlockSize} &
            \meanDiceScoreMeanFive\pm\meanDiceScoreStdFive &
            \meanDiceScoreMeanTen\pm\meanDiceScoreStdTen &
            \meanDiceScoreMeanFiveteen\pm\meanDiceScoreStdFiveteen &
            \meanDiceScoreMeanTwenty\pm\meanDiceScoreStdTwenty &
            \meanDiceScoreMeanFifty\pm\meanDiceScoreStdFifty &
            \empty,
            table foot=\bottomrule}
\end{table}
\begin{table}[ht]
    \footnotesize
    \centering
    \caption{Performance of uncertainty sampling-based strided sampling strategies on the prostate dataset.}
    \label{table:results-uncertainty-strided-sampling-prostate}
        \csvloop{
            file=data/prostate/uncertainty_strided_sampling.csv,
            head to column names,
            tabular={
                m{0.3\textwidth-2\tabcolsep}
                S[table-format=1.3(3), separate-uncertainty, table-column-width=0.135\textwidth-2\tabcolsep]
                S[table-format=1.3(3), separate-uncertainty, table-column-width=0.135\textwidth-2\tabcolsep]
                S[table-format=1.3(3), separate-uncertainty, table-column-width=0.135\textwidth-2\tabcolsep]
                S[table-format=1.3(3), separate-uncertainty, table-column-width=0.135\textwidth-2\tabcolsep]
                S[table-format=1.3(3), separate-uncertainty, table-column-width=0.135\textwidth-2\tabcolsep]
                c@{}
            },
            table head=\toprule \multirow{3}{*}{\parbox{0.45\textwidth-2\tabcolsep}{Query  Strategy}} & \multicolumn{5}{c}{Active Learning Iteration} \\ \\ & {5} & {10} & {15} & {20} & {50} \\ \midrule,
            command=
            \strategyName{\Strategy}{\Interpolation}{\BlockSize}&
            \meanDiceScoreMeanFive\pm\meanDiceScoreStdFive &
            \meanDiceScoreMeanTen\pm\meanDiceScoreStdTen &
            \meanDiceScoreMeanFiveteen\pm\meanDiceScoreStdFiveteen &
            \meanDiceScoreMeanTwenty\pm\meanDiceScoreStdTwenty &
            \meanDiceScoreMeanFifty\pm\meanDiceScoreStdFifty &
            \empty,
            table foot=\bottomrule}
\end{table}

\begin{table}[ht]
    \footnotesize
    \centering
    \caption{Quality of pseudo-labels generated through interpolation in the strided sampling strategy. The Dice scores provided in this table were calculated using the ground-truth labels.}
    \label{table:results-interpolation-quality}
        \csvloop{
            file=data/interpolation_quality.csv,
            head to column names,
            tabular={
                m{0.25\textwidth-2\tabcolsep}
                m{0.25\textwidth-2\tabcolsep}
                M{0.25\textwidth-2\tabcolsep}
                S[table-format=1.3(3), separate-uncertainty, table-column-width=0.25\textwidth-2\tabcolsep]
                c@{}
            },
            table head=\toprule {Dataset} & {Interpolation Approach} & {Max. Block Size} & {Dice Score} \\ \midrule,
            command=
            \Dataset & \Interpolation & \BlockSize & \DiceMean\pm\DiceStd &
            \empty,
            table foot=\bottomrule}
\end{table}
\FloatBarrier
\newpage

\end{document}